\documentclass[11pt, a4paper, logo, copyright, nonumbering]{aura}

\usepackage[utf8]{inputenc}
\usepackage[T1]{fontenc}
\usepackage{hyperref}
\hypersetup{
    pdfborderstyle={/S/S/W 1},
    linkbordercolor={0.4 0.7 1},
}
\usepackage{url}
\usepackage{booktabs}
\usepackage{amsfonts}
\usepackage{nicefrac}
\usepackage{microtype}
\usepackage{xcolor}
\usepackage{comment}
\usepackage{fancyvrb}
\usepackage{listings}
\usepackage{algorithm}
\usepackage{algorithmicx}
\usepackage{algpseudocode}
\usepackage{subcaption}
\usepackage{cancel}
\usepackage{xspace}
\usepackage{makecell}
\usepackage{geometry}
\usepackage{multirow}
\usepackage{blindtext}
\usepackage{multicol}
\usepackage{hyperref}
\usepackage{fontawesome5}
\usepackage{natbib}

\usepackage{microtype}
\usepackage{booktabs}
\usepackage{amsmath}
\usepackage{graphicx}
\usepackage{CJKutf8}
\usepackage[colorinlistoftodos]{todonotes}

\newcommand{\eg}{\textit{e.g.}}
\newcommand{\ie}{\textit{i.e.}}

\usepackage[most]{tcolorbox}

\newtcolorbox{MyBox}[1]{
    colback=white,
    colframe=black,
    sharp corners,dougdd
    breakable,
    left=2mm,
    right=2mm,
    top=2mm,
    bottom=2mm,
    boxrule=0.1mm,
    fontupper=\ttfamily\small
}

\definecolor{glm_bg}{RGB}{235, 245, 255}
\definecolor{glm_text}{RGB}{0, 110, 220}
\definecolor{header_gray}{RGB}{242, 242, 242}
\definecolor{section_gray}{RGB}{230, 230, 230}

\definecolor{neg}{HTML}{CB4335}
\definecolor{pos}{HTML}{27AE60}
\definecolor{delta_gray}{HTML}{6C8EBF}
\definecolor{light_bg}{RGB}{235, 245, 255}

\title{
\vspace{0.4em}
  \texorpdfstring{%
    \begin{tabular}[c]{@{}c@{\hspace{1.2em}}c@{}}
      \multirow{2}{*}[0.4em]{\includegraphics[height=3em]{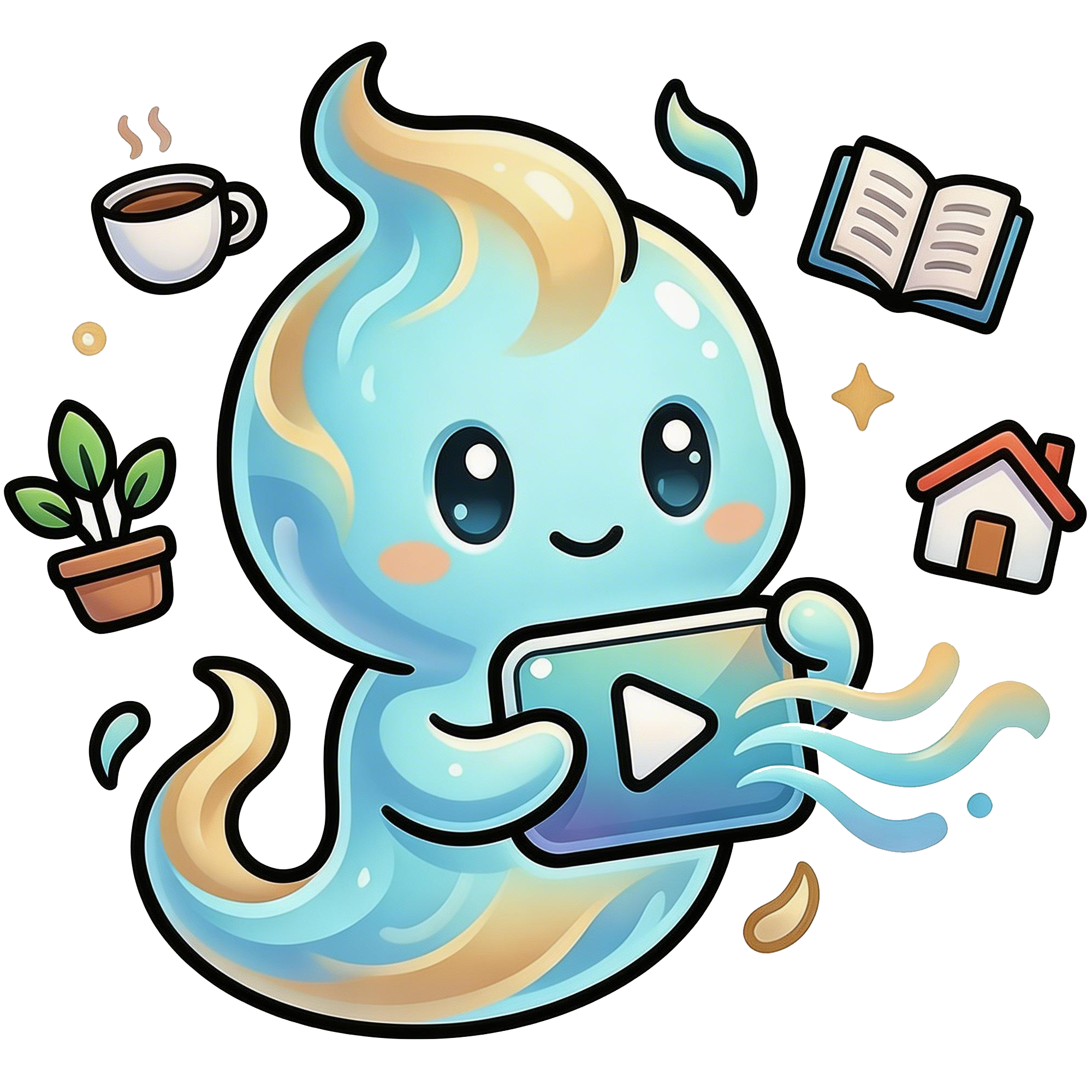}} & 
      AURA: Always-On Understanding and Real-Time \\ 
      &Assistance via Video Streams
    \end{tabular}%
  }{AURA: Always-On Understanding and Real-Time Assistance via Video Streams}%
}

\newcommand{\HFicon}{\raisebox{-0.18em}{\includegraphics[height=1.1em]{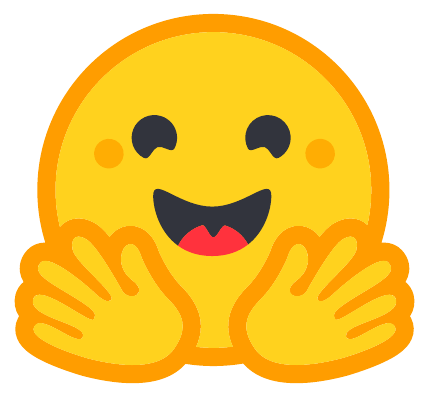}}}
\newcommand{\GHicon}{\raisebox{-0.18em}{\includegraphics[height=1.1em]{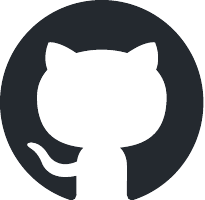}}}

\author{
    \normalsize\vspace{-0.5\baselineskip}
    {\normalfont\mdseries\small
    Xudong Lu$^{2,*}$\hspace{1em} Yang Bo$^{1,*,\dagger}$\hspace{1em} Jinpeng Chen$^{1,*}$\hspace{1em} Shuhan Li$^{1,*}$\hspace{1em} Xintong Guo$^{1,*}$\hspace{1em} Huankang Guan$^{1,*}$\newline
    Fang Liu$^{1}$\hspace{1em} Dunyuan Xu$^{1}$\hspace{1em} Peiwen Sun$^{2}$\hspace{1em} Heyang Sun$^{1}$\hspace{1em} Rui Liu$^{1,\ddagger}$\hspace{1em} Hongsheng Li$^{2,\ddagger}$}\\
    \normalsize\vspace{-0.3\baselineskip}
    {\normalfont\mdseries\small \text{$^1$Huawei Research\hspace{1.5em} $^2$CUHK MMLab}}\\
    \normalsize\vspace{0.4\baselineskip}
    {\normalfont\mdseries\footnotesize $^*$Equal Contribution \hspace{1.5em} $^\dagger$Project Lead \hspace{1.5em} $^\ddagger$Corresponding Authors}\\
    \normalsize\vspace{\baselineskip}
    \href{https://github.com/aurateam2026/AURA}{\GHicon\ \textbf{GitHub}}\hspace{1.5em}
    \href{https://huggingface.co/aurateam/AURA}{\HFicon\ \textbf{HuggingFace}}
}

\begin{document}

\begin{abstract}
Video Large Language Models (VideoLLMs) have achieved strong performance on many video understanding tasks, but most existing systems remain offline and are not well-suited for live video streams that require continuous observation and timely response. Recent streaming VideoLLMs have made progress, yet current approaches often rely on decoupled trigger-response pipelines or are limited to captioning-style narration, reducing their effectiveness for open-ended question answering and long-horizon interaction. We propose \textbf{AURA} (\textbf{A}lways-On \textbf{U}nderstanding and \textbf{R}eal-Time \textbf{A}ssistance), an end-to-end streaming visual interaction framework that enables a unified VideoLLM to continuously process video streams and support both real-time question answering and proactive responses. AURA integrates context management, data construction, training objectives, and deployment optimization for stable long-horizon streaming interaction. 
It achieves state-of-the-art performance on streaming benchmarks and supports a real-time demo system with ASR and TTS running at 2 FPS on two 80G accelerators.
We release the AURA model together with a real-time inference framework to facilitate future research.

\end{abstract}

\maketitle

\section{Introduction}\label{sec:intro}

Video Large Language Models (VideoLLMs)~\citep{zhang2023video, maaz2024video, lin2024video, ren2024timechat, li2024llava} have recently emerged as a core paradigm for video understanding by integrating visual perception with language-based reasoning and response. These models have demonstrated strong performance across a wide range of tasks, including video captioning~\citep{yang2023vid2seq,wang2024tarsier,abdar2024review}, video question answering~\citep{li2024mvbench,wu2024longvideobench,fu2025video,song2025moviechat+}, and temporal grounding~\citep{lin2023univtg,qu2024chatvtg}. They also support diverse real-world applications, such as public safety monitoring~\citep{cournan2016improving,ahn2023safefac} and intelligent transportation~\citep{wan2020intelligent,vishal2024eyes,zhang2025language}. These advances highlight the potential of VideoLLMs to become a foundational technology for next-generation intelligent visual systems.

Despite this progress, most existing VideoLLMs are still designed for offline settings~\citep{shen2024longvu,zhang2024llava,wang2025internvl3}. In this paradigm, a complete video or a pre-collected segment is first buffered and then analyzed. Although this setup is well-suited for post hoc analysis, it limits the system's ability to respond promptly to ongoing events, making it less effective for applications such as real-time AI assistants~\citep{qian2025dispider,wang2025streambridge,seed2026vision}, live video narration~\citep{chen2025livecc,xu2025streamingvlm}, and interactive robotic systems~\citep{lynch2023interactive,fang2025robix}. To address this limitation, recent studies have begun to explore streaming VideoLLMs \citep{chen2024videollm,yao2025timechat,wang2025mmduet2,seed2026vision}. Systems built on streaming VideoLLMs can continuously process incoming video frames, incrementally update their understanding of the environment, and provide real-time responses. More importantly, they move beyond the traditional paradigm in which systems respond only after receiving explicit queries, enabling proactive responses based on information discovered from the video stream.

Recent research on streaming VideoLLMs has explored effective designs in architectural design, model training, and deployment efficiency~\citep{chen2024videollm,xu2025streamingvlm,qian2025dispider,wang2025mmduet2}. Despite these advances, existing approaches still face significant challenges. Current streaming VideoLLMs mainly fall into two categories: decoupled architectures and unified architectures. Decoupled architectures~\citep{qian2025dispider, wang2025streambridge} rely on two separately deployed models, where a trigger model determines whether the primary VideoLLM should respond. Because the trigger model does not share the same contextual state with the primary model and is typically much smaller in scale, the triggering accuracy and its consistency with response generation may be limited, leading to unstable system behavior. Unified architectures offer a higher performance upper bound, but these works~\citep{chen2024videollm,xu2025streamingvlm} are generally limited to captioning-style narration tasks and remain less effective for complex open-ended video question answering. Although recent full-duplex models~\citep{minicpmo45} integrate question answering into a single framework, they still lack sufficient robustness for long-duration streaming and may suffer from memory overflow or performance degradation during extended inference.

To address these issues, we introduce \textbf{AURA} (\textbf{A}lways-On \textbf{U}nderstanding and \textbf{R}eal-Time \textbf{A}ssistance), an end-to-end streaming visual interaction framework powered by a streaming VideoLLM. Through the co-design of data, algorithms, and systems, AURA achieves high stability, broad capabilities, and long-term endurance. Specifically, AURA tackles two fundamental challenges in streaming video understanding: 1) enabling a unified model to process video streams frame by frame and autonomously decide whether to remain silent or generate an appropriate response, and 2) stably handling unbounded video-text inputs over extended durations. To this end, we propose \textbf{Interactive Video Stream Context Management}, which integrates unbounded video frame streams and textual question-answer (QA) interactions into a limited context in an appropriate manner, enabling the model to perform silent observation and support diverse response modes. Based on the streaming response modes defined by this mechanism, we further develop a \textbf{Coarse-to-Fine Data Engine} with a five-stage pipeline to systematically construct training data for Real-Time QA, Proactive QA, and Multi-Response QA. In addition, to balance continuous silence with timely responses, we introduce a \textbf{Silent-Speech Balanced Loss} during training. Furthermore, we implement a \textbf{Real-time Streaming Inference Framework}, leveraging KV-cache reuse and related optimizations to enable efficient real-time inference over unbounded video streams.

We develop AURA on top of Qwen3-VL-8B-Instruct~\citep{bai2025qwen3} as a unified streaming visual assistant for real-time and real-world interaction over continuous video streams. After training, AURA achieves state-of-the-art results on streaming benchmarks while maintaining performance comparable to that of leading models on offline benchmarks. To further support practical real-time assistance, we integrate Automatic Speech Recognition (ASR) and Text-to-Speech (TTS) into the system and develop a functional demo application. With deployment optimizations, the system runs in real time at 2 FPS on two 80G accelerators. The primary contributions of our work are summarized as follows:

\begin{itemize}

\item We study streaming visual interaction with VideoLLMs as a setting that requires continuous observation, selective silence, and timely response over unbounded video streams. We analyze the limitations of existing offline, decoupled, and unified approaches in this setting.

\item We propose AURA, an end-to-end streaming visual interaction framework that enables a unified VideoLLM to process incoming frames continuously and support both real-time question answering and proactive responses.

\item We develop a co-designed pipeline spanning data construction, training, and deployment for stable long-horizon streaming, including Interactive Video Stream Context Management for context organization, a Coarse-to-Fine Data Engine for data construction, a Silent-Speech Balanced Loss for training, and a Real-time Streaming Inference Framework for deployment.

\item AURA achieves state-of-the-art results on streaming benchmarks while maintaining competitive performance on offline video understanding tasks. To demonstrate its practical usability, we build a real-time demo system with ASR and TTS, which runs at 2 FPS on two 80G accelerators.

\end{itemize}

 \section{Related Work}\label{sec:related}

\subsection{Video Large Language Models}

Research on VideoLLMs has moved from simple feature alignment to more efficient token management and deeper reasoning in recent years. To balance spatial resolution and temporal coverage, Keye-VL-1.5~\citep{yang2025kwai} introduces a Slow-Fast video encoding strategy that dynamically allocates computation according to inter-frame similarity, enabling high-resolution modeling of key frames while preserving broader temporal coverage for relatively static content. 
Qwen3-VL~\citep{bai2025qwen3} enhances spatiotemporal modeling capabilities through interleaved-MRoPE, strengthens visual-language alignment with DeepStack, and introduces a text-based temporal alignment mechanism, thereby achieving more precise temporal localization in videos.
To reduce computational redundancy, InternVL3.5~\citep{wang2025internvl3} adopts Visual Resolution Routing and improves patch-level visual token compression without compromising performance, while LongVU~\citep{shen2024longvu} combines DINOv2~\citep{oquab2023dinov2} features with cross-modal queries to achieve adaptive spatiotemporal compression and mitigate memory bottlenecks in long-video processing. Recent studies also explore higher-level temporal reasoning and event modeling. VideoChat-A1~\citep{wang2026videochat} introduces the Chain-of-Shot reasoning paradigm to support stepwise reasoning over long videos, the VidEvent~\citep{liang2025videvent} dataset provides large-scale annotated data for modeling the evolution of dynamic events in video. Overall, these studies have advanced video modeling in terms of efficiency, reasoning, and temporal understanding, but most of them are still designed primarily for offline settings.

\subsection{Streaming Video Understanding}

The core of streaming video understanding lies in the shift from whole-video processing to online interactive processing~\citep{wang2025streambridge,xia2025streaming}. This shift requires the synchronization of perception, decision-making, and response, making streaming settings more challenging than conventional offline video understanding. Dispider~\citep{qian2025dispider} proposes a decoupled framework for perception and reaction, using a lightweight active streaming module to detect appropriate interaction points and avoid the latency introduced by autoregressive decoding during real-time monitoring. In proactive decision-making, MMDuet2~\citep{wang2025mmduet2} applies multi-round reinforcement learning and a PAUC reward mechanism to improve both response quality and timeliness without relying on precise timestamp annotations. MiniCPM-o-4.5~\citep{minicpmo45} extends omni-modal modeling to full-duplex live streaming by supporting simultaneous video and audio input together with concurrent text and speech output in an end-to-end framework. LiveCC~\citep{chen2025livecc} shows that low-cost ASR transcripts can serve as scalable supervision for streaming training, outperforming offline models of up to 72 billion parameters. StreamingVLM~\citep{xu2025streamingvlm} enables stable real-time understanding of infinite video streams through compact cache-based streaming inference and chunk-level supervised fine-tuning. VideoLLM-online~\citep{chen2024videollm} introduces a LIVE framework for streaming video dialogue, combining streaming-oriented training, dialogue data generation, and efficient real-time inference for continuous video understanding. These advances have been evaluated on benchmarks including OVO-Bench~\citep{niu2025ovo}, StreamingBench~\citep{lin2024streamingbench}, and OmniMMI~\citep{wang2025omnimmi}, which together form a more comprehensive benchmark suite for real-time interactive video understanding. Collectively, these studies highlight the importance of interaction-aware modeling to ensure both continuity and timeliness in streaming scenarios.

\section{Interactive Video Stream Context Management and QA Types}\label{sec:stream_context}

\begin{figure}[t]
    \centering
    \includegraphics[width=\textwidth]{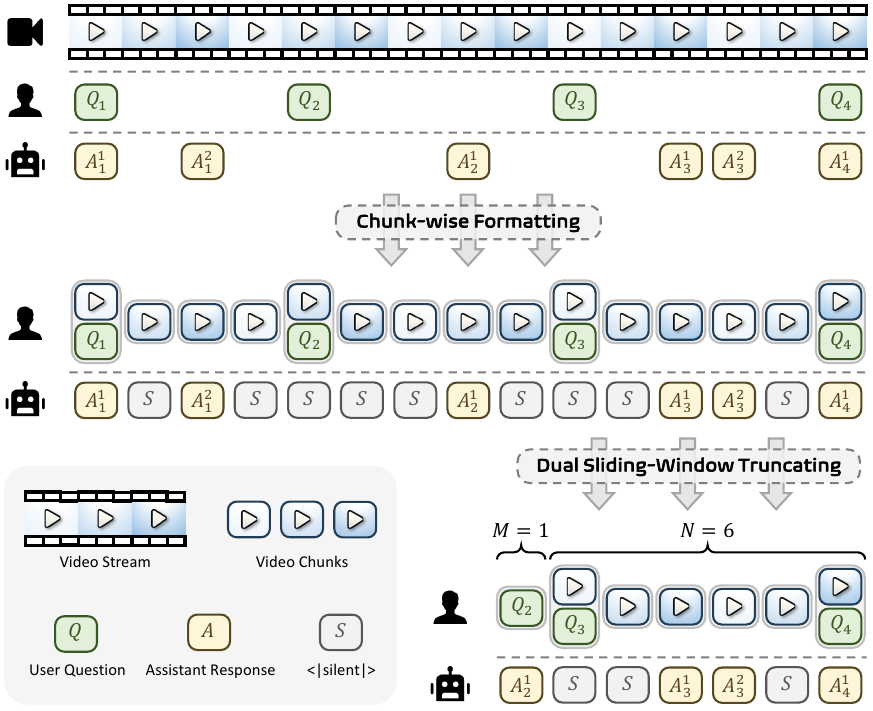}
    \caption{Overview of our Interactive Video Stream Context Management mechanism. The framework uses a dual sliding-window strategy, where $N$ denotes the length of the video window, and $M$ denotes the number of recent QA groups retained in the interaction history outside the video window.}
    \label{fig:Interactive_Video_Stream_Context_Management}
\end{figure}

Building continuous and timely visual interaction modeling for streaming scenarios is highly challenging. In streaming video understanding, both the input video and the interaction history between the user and the model (\ie, the assistant) grow continuously over time, making the continuity unbounded. However, the context window of an LLM is fundamentally limited, and an excessively long context can slow inference, thereby compromising timeliness. To address this trade-off, AURA introduces an Interactive Video Stream Context Management mechanism, as illustrated in Figure~\ref{fig:Interactive_Video_Stream_Context_Management}. Based on this mechanism, three streaming QA modes are defined.

\subsection{Interactive Video Stream Context Management}\label{subsec:interactive_video_stream_context_management}

\paragraph{Chunk-wise Conversational Format.}
Because the video stream is collected in small temporal chunks (\eg, one chunk per second), we organize the context in a chunk-wise conversational format. For each video chunk, if a user question is issued at that time, the question and the corresponding video chunk are packaged together into a user message. Otherwise, the user message contains only the video chunk and no text. Each user message is followed by an assistant message. If there is an assistant response at that time step, the assistant message contains the response text; otherwise, it is filled with a special token, \texttt{<|silent|>}, indicating that the model remains silent at that moment.

\paragraph{Dual Sliding-Window Strategy.}
To manage the unbounded growth of the streaming video and interaction history, we adopt a dual sliding-window strategy over both the video stream and the QA interaction stream. For the video stream, we maintain a sliding window that keeps only the most recent $N$ seconds of video. Because visual tokens are highly dense and user-relevant information is typically associated with recent visual content, $N$ is set to a relatively small value (\eg, $N=30$). In contrast, QA interactions are text-based, token-efficient, and often carry critical user intent and historical context. Therefore, outside the video window, we maintain a separate sliding window over QA interactions that preserves the most recent $M$ QA groups (\eg, $M=10$), where a QA group is defined as a user question together with all subsequent non-silent assistant responses. If the total video duration does not exceed $N$ seconds, the full history is retained. Otherwise, the context is truncated jointly by the video and QA windows. For the $M$ QA groups that fall outside the $N$-second video window, all video chunks and \texttt{<|silent|>} tokens are discarded, and only the valid textual content of the user and assistant messages is preserved. Overall, this dual sliding-window strategy preserves important historical information while controlling the total context length and enabling efficient computation.

\begin{figure}[t]
    \centering
    \includegraphics[width=\textwidth]{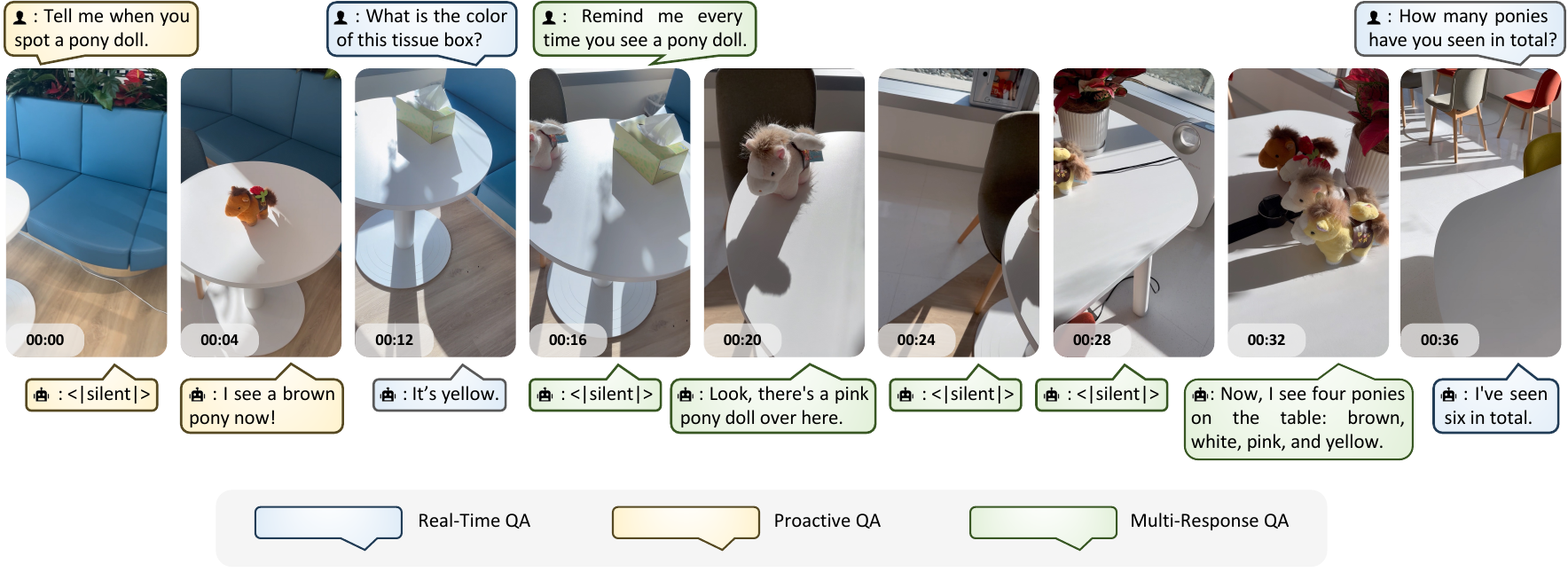}
    \caption{The figure illustrates three types of streaming QA interactions. 
\textbf{Real-Time QA} produces a single immediate response at the query time. 
\textbf{Proactive QA} produces a single delayed response after sufficient future evidence is observed. 
\textbf{Multi-Response QA} continuously tracks evolving events and produces multiple responses over time without requiring repeated queries.}
    \label{fig:QA_examples}
\end{figure}

\subsection{Streaming QA Types}

Based on the context management mechanism in Section~\ref{subsec:interactive_video_stream_context_management}, user-assistant interaction is no longer restricted to the traditional turn-taking paradigm, where a single response is generated immediately after each query. Instead, both the user and the assistant can interact asynchronously over a continuous video stream, resembling natural human conversation.
In such streaming settings, user queries exhibit diverse temporal dependencies on visual information. Some queries can be resolved immediately based on the current or previous observations, while others require waiting until sufficient evidence emerges in the future. Additionally, certain queries pertain to ongoing events, requiring the model to continuously monitor the scene and produce multiple responses as new information becomes available, without repeated user input.

These interaction patterns reflect diverse response requirements in streaming settings, varying in both response timing and frequency. Accordingly, we categorize streaming QA interactions into three types according to the timing and multiplicity of responses for each query:

\textbf{(1) Real-Time QA}, where the model produces a single immediate response grounded in the currently available or previously observed visual context;

\textbf{(2) Proactive QA}, where the model remains silent after receiving the query and generates a single response only after sufficient visual evidence has been accumulated;

\textbf{(3) Multi-Response QA}, where the model generates multiple responses over time as new visual information becomes available.

Examples of these interaction types are illustrated in Figure~\ref{fig:QA_examples}. Together, these three categories form the foundation of our data construction pipeline (Section~\ref{sec:data_pipe}), enabling the model to learn diverse response behaviors under streaming conditions.

\section{Coarse-to-Fine Streaming Data Engine}\label{sec:data_pipe}

\begin{figure}[t]
    \centering
    \includegraphics[width=\textwidth]{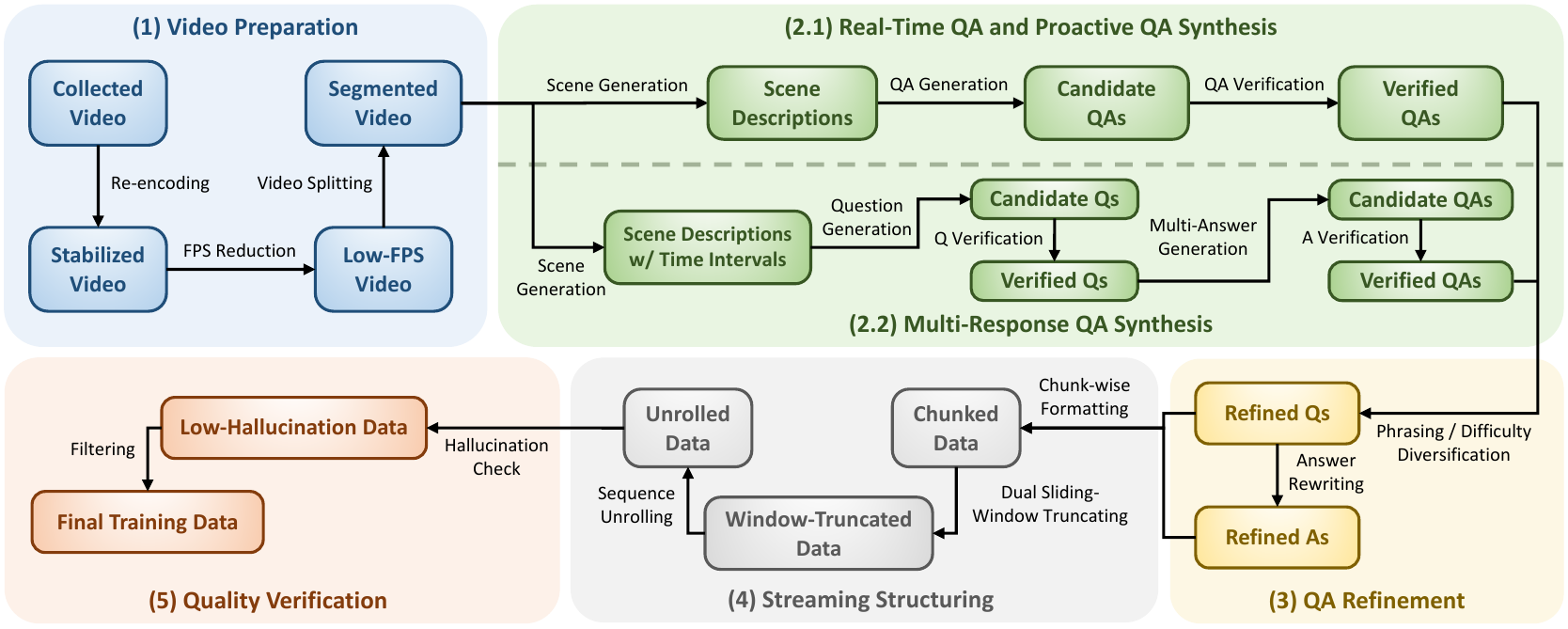}
    \caption{Overview of the Coarse-to-Fine Streaming Data Engine in AURA. The pipeline comprises five stages: (1) Video Preparation, (2) QA Synthesis, (3) QA Refinement, (4) Streaming Structuring, and (5) Quality Verification.}
    \label{fig:data_pipeline}
\end{figure}

Building on the QA categories defined above, we develop a coarse-to-fine data engine to generate streaming training data for Real-Time QA, Proactive QA, and Multi-Response QA. As illustrated in Figure~\ref{fig:data_pipeline}, the pipeline consists of five stages:  (1) \textbf{Video Preparation}, (2) \textbf{QA Synthesis}, (3) \textbf{QA Refinement}, (4) \textbf{Streaming Structuring}, and (5) \textbf{Quality Verification}.
This pipeline translates the interaction taxonomy into structured supervision, enabling the model to learn both when to respond and what to respond under different interaction patterns, thereby providing a strong data foundation for AURA.

\subsection{Video Preparation}
We collect high-quality videos from public internet sources covering a broad range of categories, including sports, vlogs, documentaries, encyclopedic content, TV shows, movies, courses, games, and animation. Then, each video is resampled to a fixed frame rate of 2 FPS to balance temporal coverage and computational cost, and re-encoded in H.264 format~\citep{wiegand2003overview} to improve codec consistency and decoding stability, while mitigating issues caused by corrupted or undecodable frames.

\subsection{QA Synthesis}
After obtaining standardized videos from the previous stage, we design two distinct pipelines to synthesize streaming QA. This separation is motivated by the different temporal structures of the QA types: Real-Time and Proactive QA require only a single response, whereas Multi-Response QA involves multiple valid answers to the same question at different time points. In the following, we describe each pipeline in detail.

The first pipeline focuses on Real-Time and Proactive QA. For each video, we first apply a multimodal large language model (MLLM) to perform scene segmentation and generate scene-level descriptions for each segment, thereby capturing the temporal progression of events throughout the video. Based on the video and its scene-level descriptions, the MLLM then produces candidate QA pairs with associated question and answer timestamps. For Real-Time QA, the question and answer share the same timestamp; for Proactive QA, the question timestamp precedes the answer timestamp. We then ask the MLLM to verify each candidate QA pair using only the video content up to the answer timestamp. For Real-Time QA, the MLLM checks whether the question is reasonable, whether the answer is correctly supported by the video content, and whether the timestamp is accurate. For Proactive QA, the MLLM further checks whether the question can be naturally raised at the question timestamp and whether sufficient information is available by the answer timestamp. Only QA pairs that pass the corresponding verification step are retained.

The second pipeline is designed for Multi-Response QA. For each video, we first apply an MLLM to perform scene segmentation and generate scene-level descriptions together with the corresponding time intervals for each scene. Then, for each video clip spanning a given time interval, we provide its scene description to the MLLM to generate candidate questions and their associated timestamps. Next, each candidate question is checked to determine whether it is reasonable and whether it can produce multiple valid answers at different timestamps within the video clip. Only the questions that satisfy both conditions are retained. For each retained question, we again prompt the MLLM with the corresponding video clip to generate multiple answers and their associated timestamps. Each generated answer is then verified to ensure that it can be correctly inferred at the specified timestamp and that it appropriately answers the question. Finally, the validated answers are retained to construct Multi-Response QA samples.

\subsection{QA Refinement}

We observe that QA samples generated in the QA Synthesis stage still exhibit limited diversity. In particular, Real-Time QA tends to exhibit limited diversity in difficulty levels, while Proactive QA and Multi-Response QA often show limited diversity in question phrasing. To better capture the diversity of difficulty levels and phrasings in real-world user queries, we design tailored refinement strategies for different QA types.

For Real-Time QA, we focus on enriching the diversity in difficulty levels. Specifically, for each original QA pair associated with a timestamp, we prompt the MLLM to generate four additional questions anchored at the same timestamp as the original question. These questions are designed to span progressively increasing levels of difficulty, from simple perceptual recognition to advanced understanding and reasoning. As a result, we obtain five candidate questions for each timestamp, including the original question and four augmented questions at different difficulty levels.
We then sample one question from these five candidates using a balanced sampling ratio. After the question is selected, we prompt the MLLM to generate an answer aligned with the sampled question based on the same video prefix, yielding the final refined QA pair.

For Proactive QA and Multi-Response QA, we focus on increasing the diversity in question phrasing by rewriting each question into different yet semantically equivalent forms. Specifically, we predefine a set of candidate templates for both Proactive and Multi-Response questions. For each question, we randomly select one template and use an LLM to generate a rewritten version accordingly. During this process, key entities, actions, and temporal references are kept unchanged to ensure consistency with both the original meaning and the video content. In this way, we enhance linguistic diversity without changing the core semantics of the original QA pairs.

\subsection{Streaming Structuring}
\label{sec:strm_struc}

Once timestamped QA annotations are obtained, this stage converts them into training samples that match the Interactive Video Stream Context Management mechanism through chunk-wise formatting and dual sliding-window truncating. In actual streaming inference, the model generates each response online based on the video window and the retained QA history available at that moment. Accordingly, during training, answers with different timestamps within the same continuous QA sequence should be paired with their corresponding video content and contextual history. We therefore unroll each sequence of continuous QA interactions from the same video into multiple training samples, each containing the interaction history up to one non-silent assistant message to be supervised, which we refer to as the target answer. For each sample, we use the timestamp of the target answer as the anchor for determining the time window. During training, we supervise only this target answer, while retaining the earlier QA interactions as contextual history (the specific supervision target is described in Section~\ref{silent-speech_balanced_loss}). This design helps bridge the gap between the necessary window truncation in long streaming contexts and the need for reliable supervision signals.

\subsection{Quality Verification}
\label{sec:quality_verify}

Since the previous stage reformats each sample within a truncated streaming context window, the retained video content and contextual history may become insufficient to support the target answer. Training on such samples may lead the model to generate content that is not grounded in the available visual input, thereby increasing the risk of hallucination. To address this issue, we introduce a dedicated Quality Verification stage.

For Real-Time QA, verification focuses on whether the target answer is supported by the relevant visual evidence available up to the response moment together with the retained QA history. The judge evaluates whether the answer is visually grounded, factually correct, temporally consistent, and free of hallucination. Only samples that satisfy these criteria are retained. For Proactive QA and Multi-Response QA, verification focuses on whether the target answer is temporally appropriate and whether its content is accurate and grounded in the retained video context and QA history. Samples are retained only when both criteria are satisfied. This stage filters out unreliable samples, helping ensure that the remaining data meet quality standards.

\subsection{Additional Details}

Beyond the five-stage pipeline, we introduce two practical designs to better align the training data with real-world streaming interactions. First, for Proactive QA and Multi-Response QA, we insert a short acknowledgment immediately after each user query to indicate that the request has been received and that an actual response will follow later, thereby improving the user experience. Second, because real streaming sessions often involve multiple interaction patterns within a single video, we combine Real-Time QA, Proactive QA, and Multi-Response QA generated from the same source video into mixed interaction sequences according to their timestamps. This design helps the model handle interleaved QA patterns more effectively in realistic streaming scenarios.

\section{Training and Inference Design}

To better address the unique challenges of streaming video understanding, we design both an effective training objective and an efficient inference framework for AURA. Specifically, we introduce a loss that supports appropriate supervision selection and mitigates silent-turn bias in streaming interactions. We also develop a real-time inference framework that enables low-latency, long-horizon assistance over continuous video streams.

\subsection{Silent-Speech Balanced Loss}\label{silent-speech_balanced_loss}

Due to the sliding-window construction of the training data, a challenge arises in deciding which assistant messages should be supervised. Because each training sample is truncated by the windowing process, not every non-silent assistant message remains fully supported by the retained visual and history evidence. As described in Section~\ref{sec:strm_struc}, during data construction, we can guarantee only that the last non-silent assistant message in each sample is paired with the video window and contextual history anchored to its timestamp, and our quality verification in Section~\ref{sec:quality_verify} is also performed with respect to this target answer. Earlier responses may have been correct in the original full sequence, but after truncation, the remaining context evidence may no longer be sufficient to justify them. Directly using all the non-silent assistant messages as supervision targets would encourage the model to produce responses from incomplete evidence, thereby increasing the risk of hallucination. We therefore apply loss only to all silent assistant messages and the last non-silent assistant message in each training sample, while excluding earlier non-silent assistant messages.

Another major challenge is the imbalance between silence and speaking. In streaming interactions, the assistant should remain silent at most timestamps and respond only when the video context or the user query truly calls for it. Consequently, in streaming training data, the special token \texttt{<|silent|>} appears much more frequently than non-silent assistant messages. Moreover, because we supervise only the last non-silent assistant message, this imbalance is further amplified in the supervision signal. If all assistant messages were optimized with a standard cross-entropy loss, the model could become overly biased toward predicting silence. To address this issue, we adopt a class-balancing strategy for assistant-side supervision. Let $N_{\text{silent}}$ denote the number of silent assistant messages in a training sample. We assign weight $1$ to target tokens from non-silent responses and down-weight target tokens from silent assistant messages by the inverse imbalance ratio, \ie, $w_{\text{silent}}=\frac{1}{N_{\text{silent}}}$, so that the two types of supervision contribute at a comparable scale. This strategy preserves the abundant silent supervision needed to learn when not to speak, while preventing it from dominating the optimization signal for informative assistant responses.

Combining supervision selection and class reweighting, the final training objective is
\begin{equation}
\mathcal{L}
=
-\frac{1}{\sum_{t=1}^{T} m_t}
\sum_{t=1}^{T}
m_t\, w_t \log p_{\theta}(y_t \mid x, y_{<t}),
\quad
w_t=
\begin{cases}
\dfrac{1}{N_{\text{silent}}}, & y_t \text{ belongs to a silent message},\\[4pt]
1, & \text{otherwise}.
\end{cases}
\end{equation}
where $T$ is the total number of target tokens in the current training instance, $y_t$ is the target token at position $t$, and $x$ denotes the multimodal streaming input. Here $m_t \in \{0,1\}$ is a supervision mask: $m_t=1$ if $y_t$ belongs to a silent assistant message or to the last non-silent assistant message in the instance, and $m_t=0$ otherwise.

\subsection{Real-Time Streaming Inference Framework}
\label{subsec:end_to_end_inference_system}

\begin{figure}[t]
    \centering
    \includegraphics[width=0.95\textwidth]{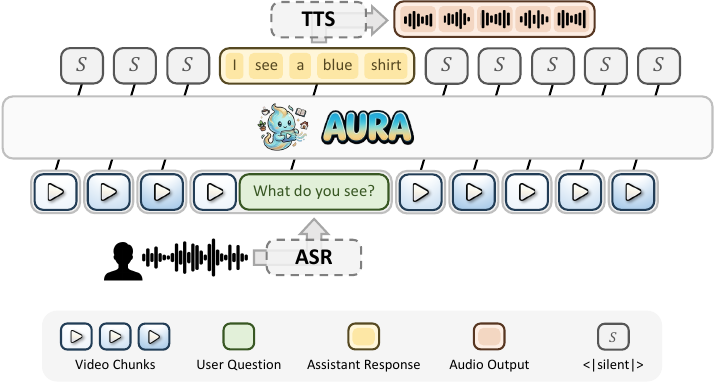}
    \caption{Overview of AURA's end-to-end real-time inference system with video and speech input, multimodal inference, and speech output. The system is designed to support continual streaming perception and low-latency interaction.}
    \label{fig:inference_pipe}
\end{figure}

To support AURA for continuous streaming perception and real-time audio-video interaction, we develop an end-to-end inference system, as illustrated in Figure~\ref{fig:inference_pipe}. The system is centered on the AURA model, integrates ASR and TTS modules, and incorporates several efficiency optimizations.

The overall context is constructed based on the Interactive Video Stream Context Management mechanism described in Section~\ref{subsec:interactive_video_stream_context_management}. On the input side, the video stream and user speech are captured simultaneously. The video stream is divided into small chunks according to a predefined chunk size. When no user speech is detected, each video chunk is independently inserted into the context as a user message. When user speech is received, it is first transcribed into text by the ASR module and then combined with the video chunk corresponding to the time at which the speech is received. The combined content is then inserted into the context as a single user message. Whenever a new user message is added to the context, the AURA model is invoked by the inference layer to generate a text response. Any non-silent response is then sent to a streaming TTS module and converted into speech, and both the text and speech are delivered to the output side. Meanwhile, the output is also added to the context as an assistant message.

Following the definition of the Interactive Video Stream Context Management mechanism, context truncation is required once the total context reaches the predefined video window size $N$. A common approach is to maintain all video chunks in the context as a fixed-length first-in-first-out (FIFO) queue, removing the oldest chunk whenever a new one is added. However, this design causes the context prefix to change continuously, which prevents the reuse of previously computed KV caches (prefix caching) and significantly reduces inference efficiency. To address this issue, we adopt an improved strategy that allows the video window to slightly extend within a margin of $N'$. Specifically, when the window size reaches $N+N'$, we remove the oldest $N'$ video chunks at once, together with the corresponding silent assistant messages. During the insertion of the next $N'$ video chunks, no further chunks are removed, which allows the KV prefix to be reused continuously. Only when the queue length again reaches $N + N'$ do we truncate the window back to $N$ by removing the oldest $N'$ video chunks, and then recompute the KV cache for the remaining $N$ chunks. For the text QA sliding window, we similarly allow it to extend slightly above $M$. However, in order to avoid increasing the frequency of KV prefix cache recomputation, we do not set a separate truncation trigger for QA. Instead, whenever the video window is truncated, we simultaneously trim the text QA groups that lie outside the video window to $M$. Since QA accumulates relatively slowly, the floating range above $M$ will also remain small. Although these two floating-window designs may cause the inference-time context length to be slightly larger than that used during training, the excess is small. In practice, we find that this design does not lead to noticeable performance degradation while significantly reducing computational cost.

Throughout the system, AURA, ASR, and TTS operate asynchronously, allowing AURA to continue running while ASR and TTS are in progress. Based on this design, we further incorporate several optimization techniques, including streaming output, computation graph optimization, and multi-process resource isolation. We also develop the frontend UI and communication protocols, resulting in a practical demo with a stable user experience that is used to evaluate AURA’s performance in real-world applications.

\section{Experiment}\label{sec:exp}

In this section, we detail the training and inference setup of AURA, report its benchmark accuracy and inference performance, and present additional experimental analyses.

\subsection{Implementation Details}
\label{subsec:training_detail}

\begin{figure}[t]
    \centering
    \includegraphics[width=0.9\textwidth]{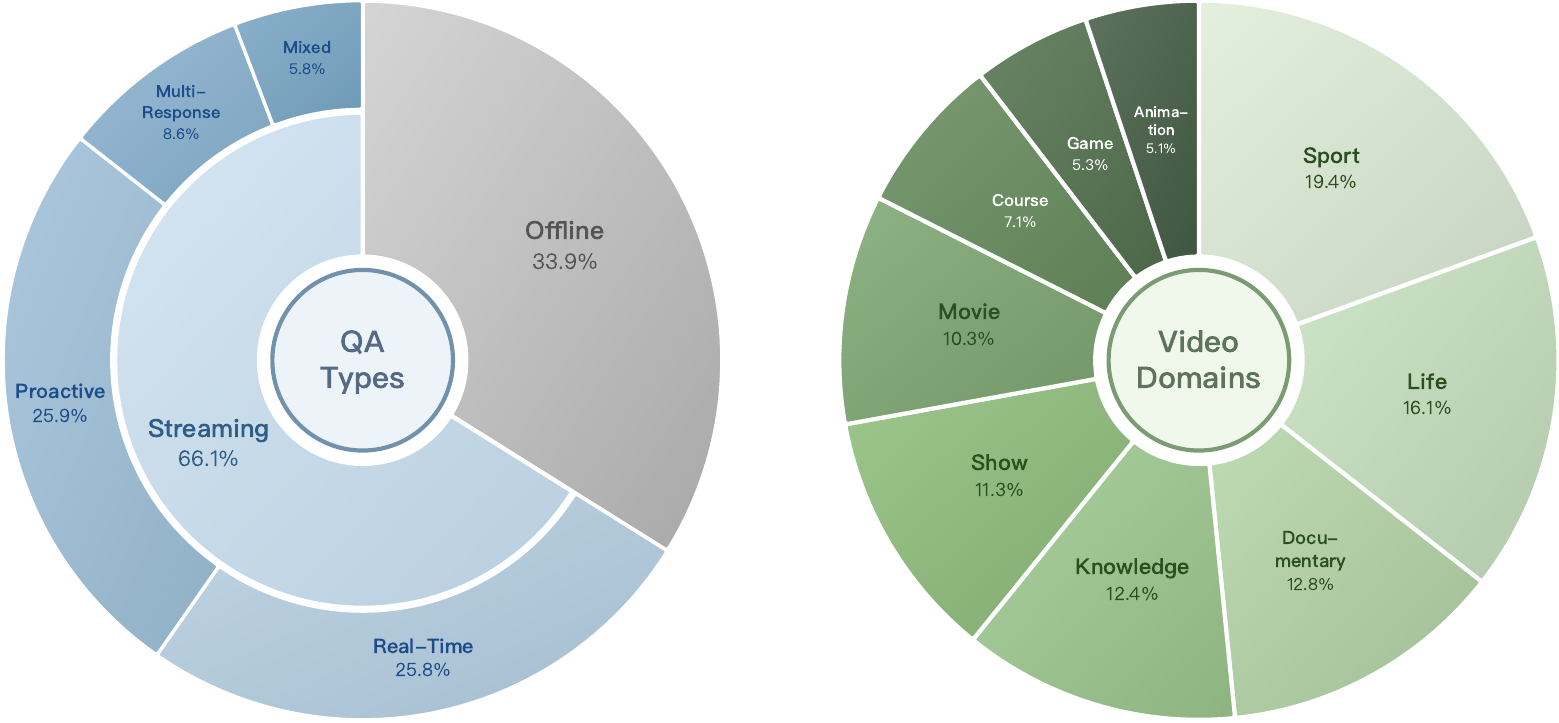}
    \caption{Training data distribution: Left: QA type distribution; Right: Video domain distribution. It shows that the training set covers diverse question formats and video domains.}
    \label{fig:data_distribution}
\end{figure}

\begin{table*}[t]
  \renewcommand{\arraystretch}{1.3}
  \caption{Quantitative comparison on StreamingBench in terms of accuracy (\%). We report performance on each fine-grained task, along with the average results for Real-Time Visual Understanding (RTVU), Omni-Source Understanding (OSU), Contextual Understanding (CU), and the overall average. The best and second-best results are highlighted in bold and underlined, respectively.}
  \label{tab:streamingbench}
  \centering
  \resizebox{\textwidth}{!}{%
  \setlength{\tabcolsep}{2pt}%
  \footnotesize
  \begin{tabular}{l|l*{21}{c}}
    \toprule
    \multirow{2}{*}{Method} & \multicolumn{11}{c}{\textbf{RTVU}} & \multicolumn{5}{c}{\textbf{OSU}} & \multicolumn{5}{c}{\textbf{CU}} & \multirow{2}{*}{\textbf{All}} \\
    \cmidrule(lr){2-12}
    \cmidrule(lr){13-17}
    \cmidrule(lr){18-22}
    & OP & CR & CS & ATP & EU & TR & PR & SU & ACP & CT & Avg & ER & SCU & SD & MA & Avg & MCU & ACU & SQA & PO & Avg & \\
    \midrule
    \multicolumn{23}{c}{\textit{Proprietary Models}} \\
    \midrule
    GPT-4o & 77.1 & 80.5 & 83.9 & 76.5 & 70.2 & 83.8 & 66.7 & 62.2 & 69.1 & 49.2 & 73.3 & 41.2 & 37.2 & 43.6 & 56.0 & 44.5 & 41.2 & 38.4 & 32.8 & 56.9 & 38.7 & 60.2 \\
    Gemini-1.5-Pro & 79.0 & 80.5 & 83.5 & 79.7 & 80.0 & 84.7 & 77.8 & 64.2 & 72.0 & 48.7 & 75.7 & 46.8 & 39.6 & 74.9 & 80.0 & 60.2 & 51.4 & 40.7 & 54.8 & 45.1 & 48.7 & 67.1 \\
    \midrule
    \multicolumn{23}{c}{\textit{Open-Source Models}} \\
    \midrule
    StreamAgent & 79.6 & 78.3 & 79.3 & 75.9 & 74.7 & 76.9 & \underline{82.9} & 66.3 & 73.7 & 55.4 & 74.3 & 35.9 & 26.3 & 38.9 & 44.0 & 36.3 & 39.7 & 30.3 & 39.6 & 28.9 & 34.6 & 57.0 \\
    ViSpeak & 79.8 & 71.1 & 81.4 & 78.8 & 74.5 & 70.1 & 63.9 & 64.2 & 71.4 & 28.0 & 70.4 & \underline{47.2} & \textbf{56.4} & \underline{61.6} & \textbf{81.2} & \underline{61.6} & \underline{49.2} & 36.4 & 39.2 & 50.8 & 43.9 & 62.6 \\
    Qwen3-VL-8B-Inst. & 80.1 & 76.6 & 81.1 & 81.1 & 77.0 & 73.2 & 80.6 & 63.4 & 68.8 & \textbf{57.0} & 74.1 & 44.4 & 27.2 & 46.4 & 49.6 & 41.9 & 30.0 & 38.4 & 38.0 & \underline{52.8} & 39.8 & 59.3 \\
    MiniCPM-o-4.5 & \underline{80.9} & \textbf{88.3} & \underline{82.0} & \underline{85.0} & \underline{78.9} & \underline{85.4} & 78.7 & \underline{67.1} & \underline{75.6} & \underline{56.0} & \underline{78.2} & 44.4 & 28.0 & 41.6 & 54.4 & 42.1 & 35.2 & \underline{44.0} & \underline{51.2} & 47.6 & \underline{44.5} & \underline{62.7} \\
    AURA (Ours) & \textbf{87.5} & \underline{84.4} & \textbf{93.4} & \textbf{89.5} & \textbf{83.2} & \textbf{87.9} & \textbf{86.1} & \textbf{76.4} & \textbf{82.4} & 47.7 & \textbf{83.2} & \textbf{56.4} & \underline{48.4} & \textbf{62.8} & \underline{80.4} & \textbf{62.0} & \textbf{54.8} & \textbf{70.8} & \textbf{57.2} & \textbf{53.2} & \textbf{59.0} & \textbf{73.1} \\
    
    \bottomrule
  \end{tabular}%
  }
\end{table*}

All experiments are conducted on a computing cluster equipped with high-performance accelerators with 80 GB of HBM3 memory. Training is performed on 4 nodes with 8 accelerators per node (32 accelerators in total). We initialize our model from Qwen3-VL-8B-Instruct~\citep{bai2025qwen3} and fine-tune only the LLM component while keeping the vision encoder and the connector frozen. The training data include approximately 115k streaming video QA samples (about 1.04B tokens), constructed using the data pipeline described in Section~\ref{sec:data_pipe}, as well as approximately 59k in-house offline video QA samples (about 0.16B tokens). In total, the training set contains approximately 174k samples, corresponding to about 1.2B tokens. Detailed data distributions are shown in Figure~\ref{fig:data_distribution}. The model is trained for one epoch with a global batch size of 128 and a learning rate of $1\times10^{-5}$. For context management, the video chunk size is set to 1 second, and the hyperparameters $N$, $N'$, and $M$ are set to 30, 15, and 10, respectively.

\subsection{Evaluation Protocol}

\paragraph{Benchmarks.}
We evaluate our AURA on three streaming video understanding benchmarks: StreamingBench~\citep{lin2024streamingbench}, OVO-Bench~\citep{niu2025ovo}, and OmniMMI~\citep{wang2025omnimmi}. StreamingBench is a comprehensive benchmark designed to evaluate streaming video understanding capabilities of VideoLLMs, focusing on real-time visual understanding, omni-source understanding, and contextual understanding through multi-timestamp question answering on continuous video streams. OVO-Bench emphasizes temporal awareness in online video understanding and evaluates models under backward tracing, real-time understanding, and forward active responding scenarios using fine-grained timestamp annotations. OmniMMI is a comprehensive multimodal interaction benchmark for streaming video contexts, which evaluates not only streaming video understanding but also proactive reasoning and multimodal interaction abilities across multiple tasks.

\paragraph{Evaluation Pipeline.}\label{Eval_pipe}

We conduct all evaluations on computing nodes with the same specifications as those used for training, and use the official codebases for the corresponding benchmarks~\citep{lin2024streamingbench,niu2025ovo,wang2025omnimmi}. We manage model context using our Interactive Video Stream Context Management mechanism. For other models, we report official results when complete results are publicly available~\citep{fu2025vispeak, wang2025omnimmi, xia2025streaming, yang2025streamagent}; otherwise, we evaluate them by strictly following the code released in the official benchmark repositories~\citep{minicpmo45,bai2025qwen3}. We also note that, although MiniCPM-o-4.5 supports a full-duplex multimodal live-streaming mode, we find that it often becomes silent in this setting and may produce irrelevant responses for video streams longer than two minutes. Therefore, we do not use this mode in our evaluation.

\subsection{Main Result}

We compare AURA with strong proprietary baselines, including GPT-4o~\citep{hurst2024gpt} and Gemini-1.5-Pro~\citep{team2024gemini}, as well as open-source baselines, including StreamAgent~\citep{yang2025streamagent}, Streamo-7B~\citep{xia2025streaming}, ViSpeak~\citep{fu2025vispeak}, M4~\citep{wang2025omnimmi}, Qwen3-VL-8B-Instruct~\citep{bai2025qwen3}, and MiniCPM-o-4.5~\citep{minicpmo45}, on three representative streaming video benchmarks. As shown in Tables~\ref{tab:streamingbench},~\ref{tab:ovobench}, and~\ref{tab:omnimmi}, AURA achieves the best overall performance on all three benchmarks, demonstrating strong capability in streaming scenarios.

\begin{table*}[t]
  \renewcommand{\arraystretch}{1.3}
  \caption{Quantitative comparison on OVO-Bench in terms of accuracy (\%). We report performance on each fine-grained task, along with the average results for Real-Time Visual Perception (RTVP), Backward Tracing (BT), Forward Active Responding (FAR), and the overall average. The best and second-best results are highlighted in bold and underlined, respectively.}
  \label{tab:ovobench}
  \centering
  \setlength{\tabcolsep}{3.5pt}%
  \footnotesize
  \begin{tabular}{l|l*{15}{c}}
    \toprule
    \multirow{2}{*}{Method} & \multicolumn{7}{c}{\textbf{RTVP}} & \multicolumn{4}{c}{\textbf{BT}} & \multicolumn{4}{c}{\textbf{FAR}} & \multirow{2}{*}{\textbf{All}} \\
    \cmidrule(lr){2-8}
    \cmidrule(lr){9-12}
    \cmidrule(lr){13-16}
    & OCR & ACR & ATR & STU & FPD & OJR & Avg & EPM & ASI & HLD & Avg & REC & SSR & CRR & Avg & \\
    \midrule
    \multicolumn{17}{c}{\textit{Proprietary Models}} \\
    \midrule
    GPT-4o & 69.8 & 64.2 & 71.6 & 51.1 & 70.3 & 59.8 & 64.5 & 57.9 & 75.7 & 48.7 & 60.8 & 73.2 & 27.6 & 59.4 & 53.4 & 59.5 \\
    Gemini-1.5-Pro & 85.9 & 67.0 & 79.3 & 58.4 & 63.4 & 62.0 & 69.3 & 58.6 & 76.4 & 52.6 & 62.5 & 35.5 & 74.2 & 61.7 & 57.2 & 63.0 \\
    \midrule
    \multicolumn{17}{c}{\textit{Open-Source Models}} \\
    \midrule
    StreamAgent & 71.2 & 53.2 & 63.6 & 53.9 & 67.3 & 58.7 & 61.3 & 54.8 & 58.1 & 25.8 & 41.7 & \underline{35.9} & 48.4 & 52.0 & 45.4 & 49.4 \\
    Streamo-7B & \underline{77.2} & \underline{66.1} & \underline{76.7} & 45.5 & 66.3 & \underline{72.8} & \underline{67.4} & 55.6 & 58.1 & 33.9 & 49.2 & 30.8 & 57.6 & \textbf{82.5} & \textbf{57.0} & 57.9 \\
    ViSpeak & 75.2 & 58.7 & 71.6 & 51.1 & \underline{74.3} & 66.9 & 66.3 & \textbf{59.9} & 48.7 & \textbf{64.0} & \underline{57.5} & 33.8 & \underline{68.5} & 60.4 & 54.3 & \underline{61.1} \\
    MiniCPM-o-4.5 & 73.2 & 65.1 & 75.0 & 59.0 & 69.3 & 62.5 & 67.3 & 55.2 & \underline{66.9} & 45.7 & 55.9 & \textbf{43.0} & 66.0 & 58.8 & \underline{55.9} & 59.7 \\
    Qwen3-VL-8B-Inst. & 54.5 & 58.7 & 68.5 & \textbf{70.3} & 68.1 & 49.5 & 61.6 & \underline{58.8} & 41.1 & 23.1 & 41.0 & 21.2 & 51.3 & 40.4 & 37.7 & 46.8 \\
    AURA (Ours) & \textbf{89.9} & \textbf{79.8} & \textbf{80.2} & \underline{70.2} & \textbf{77.2} & \textbf{81.5} & \textbf{79.8} & 54.9 & \textbf{67.6} & \underline{58.6} & \textbf{60.4} & 30.4 & \textbf{75.8} & \underline{61.3} & 55.8 & \textbf{65.3} \\
    \bottomrule
  \end{tabular}%
\end{table*}

\begin{table*}[t]
  \renewcommand{\arraystretch}{1.3}
  \caption{Quantitative comparison on OmniMMI in terms of accuracy (\%). We report performance for Dynamic State Grounding (SG), Action Prediction (AP), Multi-turn Dependency Reasoning (MD), Speaker Identification (SI), Proactive Alerting (PA), and the overall average. The best and second-best results are highlighted in bold and underlined, respectively.}
  \label{tab:omnimmi}
  \centering
  \setlength{\tabcolsep}{4pt}%
  \footnotesize
  \begin{tabular}{l|l*{11}{c}}
    \toprule
    \multirow{2}{*}{Method} & \multicolumn{4}{c}{\textbf{SG}} & \multirow{2}{*}{\textbf{AP}} & \multicolumn{4}{c}{\textbf{MD}} & \multirow{2}{*}{\textbf{SI}} & \multirow{2}{*}{\textbf{PA}} & \multirow{2}{*}{\textbf{All}} \\
    \cmidrule(lr){2-5}
    \cmidrule(lr){7-10}
    & 1st & 2nd & 3rd & Avg & & 1st & 2nd & 3rd & Avg & & & \\
    \midrule
    \multicolumn{13}{c}{\textit{Proprietary Models}} \\
    \midrule
    GPT-4o & 48.7 & 17.0 & 5.6 & 15.0 & 39.5 & 34.3 & 15.6 & 7.7 & 12.3 & 17.0 & \ding{55} & 16.8 \\
    Gemini-1.5-Pro & 52.3 & 19.7 & 9.4 & 16.3 & 43.0 & 35.0 & 16.3 & 7.1 & 12.0 & 38.5 & \ding{55} & 22.0 \\
    \midrule
    \multicolumn{13}{c}{\textit{Open-Source Models}} \\
    \midrule
    M4 & 35.7 & 6.4 & \underline{1.9} & 5.7 & \underline{33.5} & 35.7 & 6.4 & 1.9 & 1.7 & 9.0 & \underline{25.5} & 15.1 \\
    MiniCPM-o-4.5 & 46.3 & 8.1 & 0.9 & 6.3 & 25.0 & \underline{36.0} & \textbf{13.5} & \textbf{5.1} & \underline{9.3} & 17.0 & \ding{55} & 11.5 \\
    Qwen3-VL-8B-Inst. & \underline{51.3} & \underline{9.5} & 0.0 & \underline{7.7} & \textbf{36.5} & 33.7 & 11.1 & 3.5 & \textbf{9.7} & \textbf{29.5} & \ding{55} & \underline{16.7} \\
    AURA (Ours) & \textbf{52.7} & \textbf{26.4} & \textbf{15.0} & \textbf{24.0} & 32.0 & \textbf{37.3} & \underline{12.8} & \underline{3.6} & 7.7 & \underline{26.0} & \textbf{37.5} & \textbf{25.4} \\
    \bottomrule
  \end{tabular}%
\end{table*}

\paragraph{Performance Comparison.}

As reported in Table~\ref{tab:streamingbench}, AURA achieves the highest overall accuracy of 73.1\% on StreamingBench~\citep{lin2024streamingbench}, outperforming the strongest open-source baseline, MiniCPM-o-4.5, by 10.4\%. More importantly, this advantage is broad rather than concentrated on a few tasks. AURA ranks first among open-source models across all three high-level groups, \ie, Real-Time Visual Understanding (RTVU), Omni-Source Understanding (OSU), and Contextual Understanding (CU), and on 14 of the 18 fine-grained sub-tasks. AURA is also highly competitive with proprietary models, surpassing Gemini-1.5-Pro by 6.0\% overall (73.1\% vs.\ 67.1\%), while leading all proprietary models on the three group averages and on 15 of the 18 fine-grained sub-tasks.

This strong advantage generalizes to OVO-Bench~\citep{niu2025ovo}, as reported in Table~\ref{tab:ovobench}. AURA again obtains the highest overall accuracy of 65.3\%, improving over the best open-source result from ViSpeak by 4.2\%. At a finer granularity, AURA is the best open-source model on 2 of the 3 high-level settings, namely Real-Time Visual Perception (RTVP) and Backward Tracing (BT), and it trails the best model by only 1.2\% on Forward Active Responding (FAR). Compared with proprietary models, AURA remains competitive and even achieves a higher overall score than Gemini-1.5-Pro by 2.3\% (65.3\% vs.\ 63.0\%).

The same trend is also observed on OmniMMI~\citep{wang2025omnimmi}, as shown in Table~\ref{tab:omnimmi}, where AURA achieves the best overall accuracy of 25.4\%, surpassing all open-source and proprietary models. It ranks first on 5 of the 9 fine-grained metrics. It should be noted that non-streaming models (including MiniCPM-o-4.5 in non-duplex mode) do not have Proactive Alerting (PA) capability, and therefore PA is left blank.

\subsection{Inference Performance}

\paragraph{Inference Performance of AURA.}

\begin{figure}[t]
    \centering
    \includegraphics[width=\textwidth]{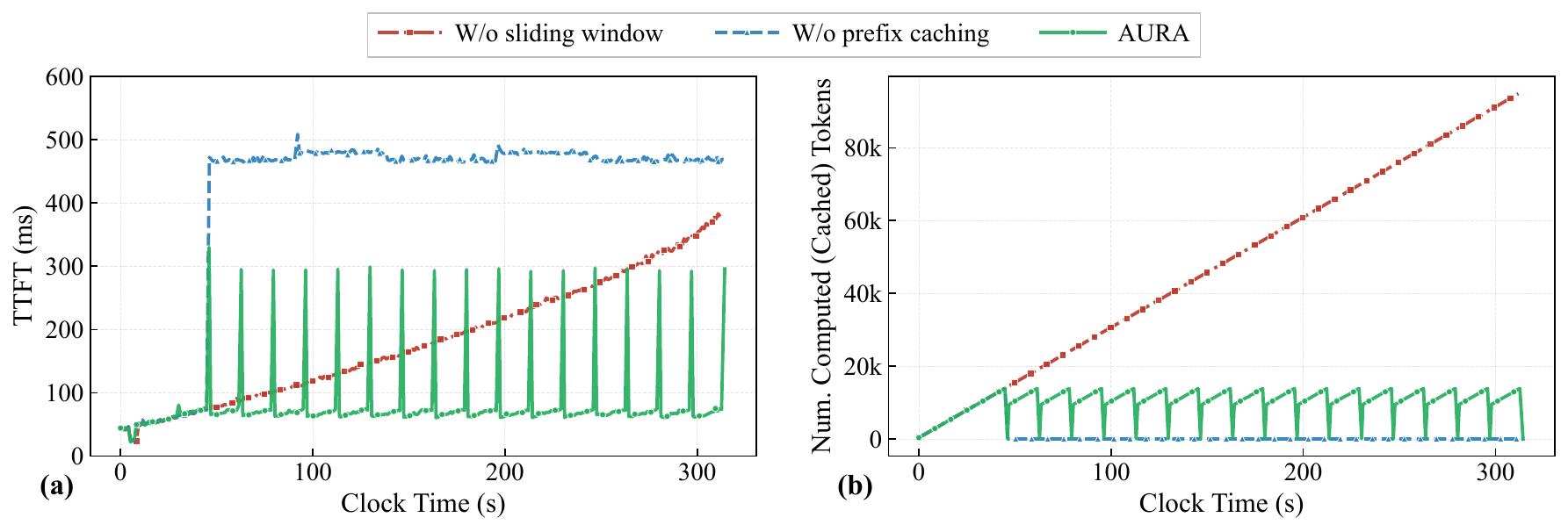}
    \caption{Inference performance of AURA. The figure compares (a) TTFT and (b) computed-token count across three settings: \textit{w/o sliding window}, \textit{w/o prefix caching}, and \textit{AURA}.}
    \label{fig:infer_pipeline}
\end{figure}

We implement the AURA inference engine on top of vLLM~\citep{vllm} and evaluate it in a latency-sensitive online video understanding setting. We use the same high-performance accelerators described in Section~\ref{subsec:training_detail} to deploy the system. We compare three configurations to isolate the contribution of our context management and cache reuse mechanisms: \textit{w/o sliding window}, which disables the sliding-window context management in Section~\ref{subsec:interactive_video_stream_context_management}; \textit{w/o prefix caching}, which disables prefix KV reuse by setting $N' = 1$; and \textit{AURA}, which enables the sliding window and uses $N' = 15$. For all settings, we use batch size 1, stream video continuously for 5 minutes at 2 FPS, and report time-to-first-token (TTFT), defined as the server-side latency between issuing a user query and receiving the first generated text token. A lower TTFT indicates better real-time responsiveness.


Figure~\ref{fig:infer_pipeline} shows that \textit{AURA} achieves the lowest average TTFT. The left panel shows that disabling prefix caching keeps TTFT consistently high because the system repeatedly recomputes long prompt prefixes, whereas removing the sliding window causes latency to increase over time as the multimodal context accumulates. This trend is corroborated by the right panel, which reports the scheduler's computed-token count: without sliding-window pruning, the amount of active cached computation grows steadily throughout the stream, while \textit{AURA} keeps it bounded and therefore sustains substantially lower TTFT. Overall, these results show that real-time responsiveness in \textit{AURA} depends on both sliding-window context management and prefix cache reuse, with the former controlling context growth and the latter avoiding frequent redundant recomputation across successive streaming queries.

\begin{table}[t]
  \renewcommand{\arraystretch}{1.2}
  \caption{End-to-end latency breakdown of AURA. ASR and TTS are averaged over 10 runs on a warmed-up engine. TTFT denotes the server-side time to the first output token of AURA, measured during a 5-minute continuous streaming session. We also report estimated first-sentence decode time. End-to-end latency is approximated as ASR + TTFT + first-sentence decode + TTS first-chunk latency.}
  \label{tab:inference_comparison}
  \centering
  \footnotesize
  \begin{tabular}{ccccc}
    \toprule
    ASR & TTFT & First-sent.\ decode & TTS first chunk & End-to-end \\
    \midrule
    84.2\,ms & 75.0\,ms & $\sim$60\,ms$^*$ & 93.0\,ms & $\sim$312.2\,ms \\
    \bottomrule
    \multicolumn{5}{l}{\scriptsize Decode speed: 7.3\,ms/tok ($\approx$137\,tok/s) \quad Avg.\ tokens/response: 12.6 \quad TTS RTF: 0.42} \\
    \multicolumn{5}{l}{\scriptsize TTFT: server-side mean over 5 mins (p50=74.6\,ms, p90=87.8\,ms, std=14.1\,ms)} \\
    \multicolumn{5}{l}{\scriptsize $^*$Estimated as $\sim$8\,tokens $\times$ 7.3\,ms/token for a typical first sentence.} \\
  \end{tabular}
\end{table}

\paragraph{End-to-end latency of AURA.}
We conduct independent latency analysis for each system component on the computing cluster described in Section~\ref{subsec:training_detail}. For inference deployment, we use two accelerators: one hosts both the ASR service (Qwen3-ASR-1.7B with vLLM as the inference backend) and the TTS service (Qwen3-TTS-12Hz-1.7B-Base with streaming synthesis), while the other hosts the main model (AURA with vLLM serving), with sliding-window attention and prefix caching enabled. For ASR, the test input consists of a 9.41-second Chinese speech command in MP3 format. For AURA, the test input consists of a 5-minute video at 2\,FPS. For TTS, the test input consists of a 40-character Chinese sentence. ASR and TTS are each evaluated independently over 10 runs on a fully warmed-up engine, whereas TTFT is measured during a 5-minute continuous streaming session to reflect steady-state performance under accumulated multi-turn context.

As shown in Table~\ref{tab:inference_comparison}, the ASR component transcribes the 9.41-second audio into text with an end-to-end latency of 84.2\,ms. For the main model, the server-side TTFT averages 75.0\,ms (p50\,=\,74.6\,ms, p90\,=\,87.8\,ms) over the 5-minute sustained streaming session, where sliding-window context management keeps the active context bounded and prefix caching avoids redundant recomputation across successive queries. During decoding, the generation speed is about 7.3\,ms per token (about 137\,tokens/s), and each response contains 12.6 generated tokens on average, corresponding to a total decoding time of about 100\,ms per response. However, because TTS operates in a sentence-level streaming fashion, audio synthesis can begin once the first sentence has been fully decoded. A typical first sentence is estimated to contain approximately 8 tokens, yielding an estimated first-sentence decode time of about 60\,ms; the remaining tokens are decoded in parallel with TTS synthesis and therefore do not contribute to the initial user-perceived latency. For streaming TTS, the first-chunk latency is highly stable at 93.0\,ms on average, with a real-time factor (RTF) of 0.42. Overall, the end-to-end latency from the user's speech input to the first spoken response is estimated to be approximately 312.2\,ms (ASR $\sim$84.2\,ms + TTFT $\sim$75\,ms + first-sentence decoding $\sim$60\,ms + TTS first chunk $\sim$93\,ms), which supports real-time conversational interaction.

\subsection{Research Question}

\paragraph{RQ1: Accuracy on Offline Video Benchmarks.}

To study how streaming-oriented training affects conventional offline video understanding, we compare AURA with its initialization model, Qwen3-VL-8B-Instruct, on three representative offline benchmarks, namely LongVideoBench~\citep{wu2024longvideobench}, MVBench~\citep{li2024mvbench}, and Video-MME~\citep{fu2025video}. We evaluate all three benchmarks with VLMEvalKit~\citep{duan2024vlmevalkit}, adhering to the original benchmark settings except that we use uniform 2 FPS sampling. This experiment is designed to answer a key question in our work: after optimizing the model for continuous streaming observation, selective silence, and delayed response in streaming scenarios, can it still preserve competitive offline understanding ability?

As shown in Table~\ref{tab:offline_rq1}, AURA achieves 58.8\% on LongVideoBench, 68.1\% on MVBench, and 65.1\% on Video-MME. Compared with its base model (Qwen3-VL-8B-Instruct~\citep{bai2025qwen3}), AURA remains particularly close on MVBench, while showing modest performance drops on LongVideoBench and Video-MME. Overall, these results indicate that streaming-oriented training enhances online interaction ability while largely preserving the model's offline video understanding capability.

\begin{table}[t]\Large
  \renewcommand{\arraystretch}{1.1}
  \caption{Offline benchmark comparison for studying the effect of streaming-oriented training on conventional offline video understanding. All results are reported in accuracy (\%).}
  \label{tab:offline_rq1}
  \footnotesize
  \centering
  \begin{tabular}{l|ccc}
    \toprule
    Method & LongVideoBench & MVBench & Video-MME \\
    \midrule
    Qwen3-VL-8B-Inst. & 61.9 & 69.0 & 68.6 \\
    AURA (Ours)  & 58.8 & 68.1 & 65.1 \\
    \bottomrule
  \end{tabular}
\end{table}

\begin{table*}[t]\Large
  \renewcommand{\arraystretch}{1.3}
  \caption{Ablation study on the training objective over OmniMMI in terms of accuracy (\%). We report averaged performance for Dynamic State Grounding (SG), Action Prediction (AP), Multi-turn Dependency Reasoning (MD), Speaker Identification (SI), Proactive Alerting (PA), and the overall average. The best and second-best results are highlighted in bold and underlined, respectively.}
  \label{tab:loss_rq2}
  \centering
  \setlength{\tabcolsep}{4pt}%
  \footnotesize
  \begin{tabular}{l|cccccc}
    \toprule
    Objective & \textbf{SG} & \textbf{AP} & \textbf{MD} & \textbf{SI} & \textbf{PA} & \textbf{All} \\
    \midrule
    Default Cross-Entropy Loss & \underline{22.7} & \underline{26.0} & \textbf{7.7} & \underline{25.5} & \underline{0.0} & \underline{16.4} \\
    Silent-Speech Balanced Loss (Ours) & \textbf{24.0} & \textbf{32.0} & \textbf{7.7} & \textbf{26.0} & \textbf{37.5} & \textbf{25.4} \\
    \bottomrule
  \end{tabular}
\end{table*}

\paragraph{RQ2: Effect of the Training Objective.}
To evaluate the training objective in Section~\ref{silent-speech_balanced_loss}, we conduct an ablation study on OmniMMI. Specifically, we train a variant of AURA with the same data, initialization, and training settings, but replace our objective with the default cross-entropy loss that uniformly optimizes all assistant messages. As shown in Table~\ref{tab:loss_rq2}, replacing our objective with the default loss substantially hurts overall performance: the overall average drops from 25.4\% to 16.4\%, and PA falls from 37.5\% to 0.0\%. Clear drops are also observed on State Grounding (SG), Action Prediction (AP), and Speaker Identification (SI).

Closer inspection shows that the model trained with the default loss tends to over-generate \texttt{<|silent|>}, remaining silent at every time step in PA. For non-proactive interaction metrics, the degradation is also consistent with our supervision-selection design: under sliding-window truncation, earlier non-silent assistant responses may no longer be fully supported by the preserved visual and history context, so uniformly supervising all assistant messages can introduce insufficiently grounded targets and increase hallucination risk. 
These results directly reflect the problems that our objective is designed to address: it reweights silent messages to prevent excessive silence and supervises only the final non-silent assistant message to reduce hallucinations.

\section{Conclusion}\label{sec:conclu}

In this work, we present \textbf{AURA}, an end-to-end streaming visual interaction framework that enables a unified VideoLLM to continuously observe live video streams and support both real-time question answering and proactive responses. To address the core challenges of streaming video understanding, namely selective silence, timely response, and long-horizon context management, we co-design the framework from data construction, model training, and system deployment. Specifically, AURA introduces Interactive Video Stream Context Management to organize unbounded video and interaction history, a Coarse-to-Fine Streaming Data Engine to construct supervision for Real-Time QA, Proactive QA, and Multi-Response QA, a Silent-Speech Balanced Loss to avoid excessive silent prediction during training, and an efficient Real-Time Streaming Inference Framework for stable deployment.
Extensive experiments show that AURA achieves state-of-the-art performance on streaming benchmarks while maintaining competitive performance on conventional offline video understanding tasks. Beyond benchmark evaluation, we further demonstrate the practical usability of AURA through a real-time demo system with ASR and TTS, running at 2 FPS on two 80G accelerators. Overall, our results suggest that unified streaming VideoLLMs can move beyond offline post hoc analysis toward continuous, interactive, and proactive visual assistance. We hope AURA, together with the released model and real-time inference framework, can provide a strong foundation for future research on streaming video understanding and real-world always-on visual assistants.

\bibliographystyle{unsrt}
{
\linespread{0.75}\selectfont
\bibliography{references_formated.bib}
}

\end{document}